\documentclass[copyright,hidelinks]{eptcs}
\providecommand{\event}{ICLP 2019} 
\usepackage{underscore}           

\usepackage{url}
\usepackage{hyperref}

\usepackage[T1]{fontenc}
\usepackage[utf8]{inputenc}
\usepackage{amsmath,amssymb}
\usepackage{csquotes}
\usepackage{subcaption}
\usepackage{cite}

\usepackage{graphicx}
\graphicspath{{./figures/}}

\usepackage{cleveref}
\crefname{figure}{figure}{figures}
\Crefname{figure}{Figure}{Figures}
\crefname{equation}{}{}
\Crefname{equation}{}{}

\usepackage{mathtools}
\newtagform{angles}[\textit]{$\langle$}{$\rangle$}
\usetagform{angles}
\creflabelformat{equation}{#2$\langle$\textit{#1}$\rangle$#3}

\newcommand{\mi}[1]{\ensuremath{\mathit{#1}}}

\newcommand{\slv}[1]{\textsc{#1}} 
\newcommand{\alphaslv}{\slv{Alpha}}

\newcommand{\as}{\ensuremath{\mathrm{AS}}}
\newcommand{\prog}{\ensuremath{P}}
\newcommand{\assignment}{\ensuremath{A}}
\newcommand{\assignmentpm}{\ensuremath{A}^\pm}
\newcommand{\nafsymbol}{\ensuremath{\mathrm{not}}}
\newcommand{\naf}[1]{\ensuremath{\nafsymbol~#1}}
\newcommand{\head}{\ensuremath{\mathrm{H}}}
\newcommand{\body}{\ensuremath{\mathrm{B}}}
\newcommand{\bodyp}{\ensuremath{\mathrm{B}^+}}
\newcommand{\bodyn}{\ensuremath{\mathrm{B}^-}}

\newcommand{\sigT}{\ensuremath{\mathbf{T}}}
\newcommand{\sigF}{\ensuremath{\mathbf{F}}}
\newcommand{\sigM}{\ensuremath{\mathbf{M}}}

\newcommand{\condsymbol}{\ensuremath{\mathrm{c}}}
\newcommand{\cond}[1]{\ensuremath{\condsymbol(#1)}}
\newcommand{\condpos}[1]{\ensuremath{\condsymbol^+(#1)}}
\newcommand{\condneg}[1]{\ensuremath{\condsymbol^-(#1)}}
\newcommand{\heudirstmt}{\mathtt{\#heuristic}}
\newcommand{\sumagg}{\mathtt{\#sum}}
\newcommand{\heusign}{\ensuremath{s}}
\newcommand{\heuat}{\ensuremath{\mi{ha}}}
\newcommand{\fheuat}{\ensuremath{\mathrm{atm}}}
\newcommand{\heusignempty}{\ensuremath{\epsilon}}
\newcommand{\heudir}{\ensuremath{d}}
\newcommand{\heudirs}{\ensuremath{D}}
\newcommand{\heupred}{\ensuremath{\mathrm{\_h}}}

\newtheorem{definition}{Definition}
\newtheorem{example}{Example}

\newcommand{\mytitle}[0]{Exploiting Partial Knowledge in Declarative Domain-Specific Heuristics for ASP}

\title{\mytitle}

\author
{
    Richard Taupe\textsuperscript{1,2}, Konstantin Schekotihin\textsuperscript{2}, Peter Schüller\textsuperscript{3},\\ Antonius Weinzierl\textsuperscript{3,4}, and Gerhard Friedrich\textsuperscript{2}
    \institute{
    \textsuperscript{1} Siemens AG Österreich, Corporate Technology, Vienna, Austria,
            \email{richard.taupe@siemens.com}\\
    \textsuperscript{2} Alpen-Adria-Universität, Klagenfurt, Austria,
            \email{\{konstantin.schekotihin,gerhard.friedrich\}@aau.at}\\
    \textsuperscript{3} Technische Universität Wien, Institut für Logic and Computation, KBS Group,
            \email{\{ps,weinzierl\}@kr.tuwien.ac.at}\\
    \textsuperscript{4} Aalto University, Department of Computer Science
	}
}

\def\titlerunning{\mytitle}
\def\authorrunning{R.\ Taupe, K.\ Schekotihin, P.\ Schüller, \\ ~A.\ Weinzierl \& G.\ Friedrich}

\begin{document}
\label{firstpage}

\maketitle

\begin{abstract}
Domain-specific heuristics are an important technique for solving
combinatorial problems efficiently.
We propose a novel semantics
for declarative specifications of domain-specific 
heuristics
in Answer Set Programming (ASP).
Decision procedures that are based on a partial solution are a frequent ingredient
of existing domain-specific heuristics,
e.g., for placing an item that has not been placed yet in bin packing.
Therefore, in our novel semantics
negation as failure and aggregates in heuristic conditions are
evaluated on a partial solver state.
State-of-the-art solvers do not allow
such a declarative specification.
Our implementation
in the lazy-grounding ASP system Alpha
supports heuristic directives under this semantics.
By that, we also provide the first implementation for incorporating declaratively specified domain-specific heuristics in a lazy-grounding setting.
Experiments confirm that the combination
of ASP solving with lazy grounding
and our novel heuristics can be a vital ingredient
for solving industrial-size problems.
\end{abstract}


\section{Introduction}
Answer Set Programming (ASP) \cite{aspbook-gelfond,aspbook-baral,Gebser.2012}
is a declarative knowledge representation formalism
that has been applied successfully in a variety of industrial and scientific applications
such as configuration \cite{DBLP:conf/cpaior/AschingerDFGJRT11,Hotz.2014b},
team building \cite{DBLP:journals/tplp/RiccaGAMLIL12},
molecular biology \cite{DBLP:journals/tplp/DurzinskyMOSW11},
planning \cite{DBLP:journals/ai/EiterFLPP03}, 
and others \cite{DBLP:journals/aim/ErdemGL16,DBLP:journals/ki/FalknerFSTT18}.
In the vast majority of these applications ASP solvers such as  \slv{clingo} \cite{DBLP:journals/corr/GebserKKS14} or \slv{DLV} \cite{DBLP:journals/ki/AdrianACCDFFLMP18,DBLP:journals/tocl/LeonePFEGPS06} applied the ground-and-solve approach \cite{DBLP:conf/ijcai/GebserLMPRS18}.
Such solvers first instantiate the given non-ground program and then apply various strategies to find answer sets of the obtained ground program.

Modern applications, however, showed two issues with the ground-and-solve approach.
First, problem instances in industrial applications often cannot be grounded by modern grounders like \slv{gringo} \cite{DBLP:conf/lpnmr/GebserKKS11} or \slv{I-DLV} \cite{DBLP:journals/ia/CalimeriFPZ17} in acceptable time and/or space \cite{DBLP:journals/amai/EiterFFW07}.
This issue is known as the \emph{grounding bottleneck}.
Second, even if the problem can be grounded, computation of answer sets might take considerable time, as indicated by the results of the last ASP Competitions \cite{DBLP:journals/ai/CalimeriGMR16,DBLP:journals/jair/GebserMR17,aspcomp2017}.

To overcome the grounding bottleneck, lazy-grounding ASP systems such as
\slv{gasp} \cite{DBLP:journals/fuin/PaluDPR09}, \slv{ASPeRiX} \cite{DBLP:journals/tplp/LefevreBSG17}, \slv{OMiGA} \cite{DBLP:conf/jelia/Dao-TranEFWW12}, or \alphaslv\ \cite{DBLP:conf/lpnmr/Weinzierl17}
interleave grounding and solving in order to instantiate and store only relevant parts of the ground program in memory.

To overcome the runtime performance issue modern solvers employ various techniques.
Among them the ability to use domain-specific heuristics is fundamental to solve complex problems \cite{DBLP:conf/lpnmr/GebserRS15}.
In one approach \cite{DBLP:conf/aaai/GebserKROSW13} heuristics are specified using a dedicated declarative language as a part of the encoding.
The solver then evaluates all heuristic rules as a part of the program.
Another approach \cite{DBLP:journals/tplp/DodaroGLMRS16} allows for specification of procedural heuristics that interact directly with the internal decision-making procedures and therefore can dynamically evaluate heuristics wrt a partial solution.
For example, a heuristics for bin packing may need to compute the amount of space left in a bin after an item is placed into it.

These existing approaches to integrate domain-specific heuristics with ASP solving are far from optimal:
The existing declarative approach does not permit dynamic heuristics reasoning about partial solutions,
and procedural heuristics counteract the declarative nature of ASP.
Declarative heuristics for the lazy-grounding case have not been addressed yet at all.

Finding a satisfying solution is challenging:
The existing declarative approach shall be extended s.t.\ it becomes possible to encode dynamic heuristics, while at the same time keeping the language simple and easy to use.
Integration into a lazy-grounding system requires non-trivial adaptations due to the different solving mechanisms in effect.

In this work we present a novel approach to dynamic declarative domain-specific heuristics for ASP and combine it with lazy grounding to facilitate solving of large and complex problems.
In summary, our work makes the following contributions:
\begin{itemize}
	\item we present a novel semantics that makes declarative specifications of domain-specific heuristics more intuitive, using a language which can be seen as a variant of \cite{DBLP:conf/aaai/GebserKROSW13};
	\item we show how the language can be integrated into a well-known lazy-grounding ASP system, \alphaslv, and provide a reference implementation;
	\item finally, we demonstrate the benefits of our approach with experimental results.
\end{itemize}

After briefly describing ASP syntax and semantics in \cref{sec-preliminaries}, we discuss the state of the art of domain-specific heuristics in ASP in \cref{sec-soa}.
Then, we present a novel semantics for such heuristics in \cref{sec-novel-semantics} and show how to integrate it into a lazy-grounding ASP solver in \cref{sec-integration}.
Finally, experimental results are presented and discussed in \cref{sec-experiments}.

\section{Preliminaries}
\label{sec-preliminaries}

Answer Set Programming (ASP) \cite{aspbook-gelfond,aspbook-baral,Gebser.2012} is an approach to declarative programming.
Instead of stating how to solve a problem, the programmer formulates the problem in the form of a logic program.
An ASP solver then finds models (so-called \textit{answer sets}) for this logic program, which correspond to solutions for the original problem.

\label{sec-preliminaries-syntax}

An answer-set program \prog\ is a finite set of rules of the form
\begin{equation}
  \label{eqRule}
	h \leftarrow b_1,~\ldots~,b_m,~\naf{b_{m+1}},~\ldots,~\naf{b_n}.
\end{equation}
where $h$ and $b_1,\dots,b_m$ are positive literals (i.e., atoms) and $\naf{b_{m+1}},\ldots,\naf{b_n}$ are negative literals.
An atom is either a classical atom, a cardinality atom, or an aggregate atom. A classical atom is an expression $p(t_1,\dots,t_n)$ where $p$ is an $n$-ary predicate and $t_1,\dots,t_n$ are terms.
A literal is either an atom $a$ or its default negation $\naf{a}$.
Default negation refers to the absence of information, i.e., an atom is assumed to be false as long as it is not proven to be true.
A \textit{cardinality atom} is of the form
$
l ~ \{a_1:l_{1_1},\dots,l_{1_m};\ldots; a_n:l_{n_1},\dots,l_{n_o}\} ~ u
$,
where
$a_i:l_{i_1},\dots,l_{i_m}$ represent \textit{conditional literals} in which $a_i$ (the head of the conditional literal) and all $l_{i_j}$ are literals, and
$l$ and $u$ are integer terms indicating lower and upper bound. If one or both of the bounds are not given, their defaults are used, which are $0$ for $l$ and $\infty$ for $u$.
As an extension of cardinality atoms, ASP also supports aggregate atoms that apply aggregate functions like $\mi{max}$, $\mi{min}$ or $\mi{sum}$ to such sets \cite{aspcore2}.
Given a rule $r$, $\head(r)=\{h\}$ is called the \textit{head} of $r$, and
$\body(r) = \{ b_1,\dots,b_m,$ $\naf{b_{m+1}},$
$\ldots,\naf{b_n} \}$ is called the body of $r$.
A rule $r$ with $\head(r)$ consisting of a cardinality atom is called \textit{choice rule}.
A rule $r$ where $\head(r) = \{\}$, e.g., $\leftarrow b.$, is called \textit{integrity constraint}, or simply \textit{constraint}.
A rule $r$ where $\body(r) = \{\}$, e.g., $h \leftarrow.$, is called \textit{fact}.

A (partial) \emph{assignment} $\assignment$ is a set of signed literals over $\sigT$ (\emph{true}), $\sigF$ (\emph{false}), and $\sigM$ (\emph{must-be-true}).
Its \emph{signed atom projection} $\assignmentpm := \{ -a \mid \sigF a \in \assignment \} \cup \{ a \mid \sigM a \in \assignment ~\mathrm{or}~ \sigT a \in \assignment \} \cup \{ +a \mid \sigT a \in \assignment \}$ maps an assignment to a set of atoms with corresponding $+$/$-$ sign symbols, which will be used in \cref{sec-novel-semantics} to define our novel semantics.

There are several ways to define the semantics of an answer-set program, i.e., to define the set of answer sets $\as(\prog)$ of an answer-set program \prog.
An overview is provided by \cite{DBLP:conf/birthday/Lifschitz10}.
Probably the most popular semantics is based on the \textit{Gelfond-Lifschitz reduct} \cite{DBLP:conf/iclp/GelfondL88}.
The \textit{FLP semantics} also covers aggregates \cite{DBLP:journals/ai/FaberPL11}.
A variant that applies to choice rules also is presented by \cite{DBLP:journals/ai/CalimeriGMR16}.

Informally, an answer set $A$ of a program $\prog$ is a subset-minimal model of $\prog$ (i.e., a set of atoms interpreted as \textit{true}) which satisfies the following conditions:
All rules in $\prog$ are satisfied by $A$;
and all atoms in $A$ are \enquote{derivable} by rules in $\prog$.
A rule is satisfied if its head is satisfied or its body is not.
A cardinality atom is satisfied if $l \leq |C| \leq u$ holds, where $C$ is the set of head atoms in the cardinality atom which are satisfied together with their conditions (e.g., $l_{i_1},\dots,l_{i_m}$ for $a_i$).
An aggregate atom is satisfied if the value computed by the aggregate function respects the given bounds,
e.g., $1 = \sumagg \{ 1 : \mathrm{a}; 2 : \mathrm{b} \}$ is satisfied if $\mathrm{a}$ but not $\mathrm{b}$ is true.

\section{Domain-Specific Heuristics in ASP: State of the Art}
\label{sec-soa}

State-of-the-art ASP solvers are well suited to solve a wide range of problems, as shown in ASP competitions, experiments, and (industrial) applications reported in the literature \cite{aspcomp2017,DBLP:journals/aim/ErdemGL16,DBLP:journals/ki/FalknerFSTT18,DBLP:conf/rweb/LeoneR15}.
However, applying general ASP solvers to large instances of industrial problems often does not perform well enough.
In some cases, sophisticated encodings or solver tuning methods, e.g., portfolio solvers like \slv{claspfolio} \cite{DBLP:journals/tplp/HoosLS14} or \slv{me-asp} \cite{DBLP:conf/aiia/MarateaPR15}, can significantly improve performance.

However, the major breakthrough in solving industrial configuration problems was achieved by applying domain-specific heuristics to ASP.
There is a number of approaches to embed heuristic knowledge into the ASP solving process.
\slv{hwasp} \cite{DBLP:journals/tplp/DodaroGLMRS16} extends the \slv{wasp} solver to facilitate external heuristics implemented in a procedural language which are consulted at specific points during the solving process via an API.
\slv{hwasp} could find solutions for all available instances of the Partner Units Problem (PUP) using a number of externally embedded heuristics.

One of the first approaches integrating domain-specific heuristics in ASP was suggested in \cite{DBLP:conf/aaai/GebserKROSW13}.
It extends the ASP language to allow for declarative specification of atom weights and signs for the internal heuristics of the \slv{clasp} solver.
The \slv{clingo} system supports $\heudirstmt$ directives, which are described in detail in \cite[section 10]{potasscoguide}.
The weight of an atom influences the order in which atoms are considered by the solver when making a decision and a sign modifier instructs whether the selected atom must be assigned true or false.

In \slv{clingo}, the following (non-ground) meta statement defines domain-specific heuristics, where $A$ is an atom, $B$ is a rule body, and $w$, $p$, and $m$ are terms \cite{potasscoguide}.
\begin{align}
\label{eq-clingo-heuristic-directive}
  &\heudirstmt ~~ A : B. \qquad [w@p,m]&
\end{align}
The optional term $p$ gives a preference between heuristic values for the same atom (preferring those with higher $p$).
The term $m$ specifies the type of heuristic information:
$m{=}$\texttt{true} specifies that $A$ should be guessed to true with weight $w$ if $B$ is true,
$m{=}$\texttt{false} is the analog heuristics for false.
The weight defines a partial order over atoms;
atoms with a higher weight are assigned a value before
atoms with a lower weight.
Further values for $m$ are \texttt{init} and \texttt{factor}, which allow to replace initial scores and dynamically modify scores assigned to atoms by the VSIDS heuristics underlying the domain-specific heuristics \cite{DBLP:conf/aaai/GebserKROSW13,potasscoguide}.

To the best of our knowledge, the first account of domain-specific heuristics in lazy-grounding ASP solving is presented in \cite{alpha-heuristics-paoasp}, where heuristic decisions are made procedurally.

Although purely declarative approaches are preferable to those resorting to procedural means, so far declarative approaches suffer from a modelling issue that we illustrate in the following.

\begin{example}
Consider the following program containing two heuristic directives:
\begin{align*}
&\{ ~ \mathrm{a(2)~;~a(4)~;~a(6)~;~a(8)~;~a(5)} ~ \} \leftarrow . \\
&\leftarrow \sumagg ~ \{ ~ X : \mathrm{a}(X) ~ \} = S, ~~ S \backslash 2 \neq 0.
&\%~S~\mathrm{mod}~2 \neq 0 \\
&\heudirstmt ~~ \mathrm{a(5)}.   &[1, \mathrm{true}] \\
&\heudirstmt ~~ \mathrm{a(4)} ~~ : ~~ \naf{\mathrm{a(5)}}. &[2, \mathrm{true}]
\end{align*}
The program guesses a subset of $\{ 2, 4, 6, 8, 5 \}$,
the sum of which must be even, i.e., $\mathrm{a(5)}$ must not be chosen.
The heuristic statements specify
that $\mathrm{a(5)}$ shall be set to true with weight 1,
and that $\mathrm{a(4)}$ shall be set to true if $\naf{\mathrm{a(5)}}$ is true with weight 2.
\end{example}

In solving this program, \slv{clingo} (v.\ 5.3) first assigns $\mathrm{a(5)}$ to true in our experiments,
although $\mathrm{a(4)}$ has a higher weight
and $\mathrm{a(5)}$ is not known to be true in the beginning.
Next,
$\mathrm{a(8)}$ is chosen to be false,\footnote{Choices not determined by the heuristic directives may vary from one implementation to another.}
the solver backtracks and only $\mathrm{a(5)}$ stays assigned.
Finally, $\mathrm{a(8)}$ is chosen to be true and
a conflict is learned such that after backtracking $\mathrm{a(5)}$ is set to false.
Now that $\naf{\mathrm{a(5)}}$ is satisfied,
the second heuristics chooses $\mathrm{a(4)}$ to be true,
and we obtain the answer set $\{ \mathrm{a(4), a(8)} \}$
after a few more guesses on the yet unassigned atoms.
The second heuristics becomes active only late
because $\naf{\mathrm{a(5)}}$ is evaluated
as true only if $\mathrm{a(5)}$ is false.
This has small implications in our toy example
but might be crucial for industrial problems
where heuristics must be defined depending on 
the variables that were not assigned so far by the solver.

To overcome this issue,
we propose in the following section to evaluate negation as failure
(i.e., $\nafsymbol$) in heuristic statements
\emph{with respect to the partial assignment} in the solver.
As a consequence, $\naf{X}$ is true if $X$ is false or unassigned during the search.

\begin{example}
\label{ex-bpp-clingo}
Consider the \emph{best-fit} heuristics for the \emph{Bin-Packing Problem} (BPP) that suggests to place items s.t.\ after placement the remaining space in the bin is minimal.
For a simple BPP, where sizes of items correspond to their numbers, this heuristics can be encoded as follows:
\begin{align*}
&1\{ ~ \mathrm{in}(I,B): \mathrm{bin}(B) ~ \}1 \gets \mathrm{item}(I). \\
&\gets \sumagg ~ \{ ~ I : \mathrm{in}(I,B) ~ \} > C, \mathrm{bcap}(C), \mathrm{bin}(B). \\
&\heudirstmt ~ \mathrm{in}(I,B) : \mathrm{bin}(B), \mathrm{item}(I), \mathrm{bcap}(C),\\
&~~~C \geq S+I, S = \sumagg ~ \{ ~ I' \,{:} ~ \mathrm{in}(I',B) ~ \}. ~ [S+I, \mathrm{true}] 
\end{align*}
The program guesses assignments of items to bins ($\mathrm{in/2}$) and forbids guesses in which the sum of item sizes assigned to one bin is greater than the capacity ($\mathrm{bcap/1}$).
The heuristic directive assigns weight and sign values to atoms over $\mathrm{in/2}$ and thus aims to influence choices made by the solver.
For item $I$ and bin $B$, the total size of all items already in $B$ plus the size of $I$ is computed.
If this sum is greater than the bin capacity $C$, then no weight and sign are assigned to an atom in the head and, therefore, such atoms will have the smallest priority among all choice atoms.
Otherwise, the larger the sum, the more preferred is the assignment.
In the following exemplary BPP instance there are three bins with capacity 5 and five items:
\[\mathrm{bcap(5). \qquad bin(1..3). \qquad item(1..5).}\] 
Here, according to the heuristics the solver should place item 5 first, since the sum of items in a bin after placing this item is 5 and, consequently, the remaining space 0.
Next, the heuristics would suggest to place item 4 into some bin,\footnote{Note that this depends on whether the solver can recognize at this point that an item can be placed into only one bin. If this is not the case, an additional literal $\naf{\mathrm{in}(I,\_)}$ can be used in the condition of the heuristic directive to prevent the heuristics from preferring to assign item 5 to a second bin after placing it in the first one.} followed by item 1 in the same bin, and so forth.
\end{example}

However, evaluation of aggregate atoms using the current semantics of heuristic directives does not allow this expected evaluation of the best-fit heuristics,
because this would require aggregates to be evaluated to different values on different partial assignments,
which is impossible given the semantics of aggregates \cite{DBLP:journals/tocl/Ferraris11,DBLP:journals/ai/FaberPL11} implemented in \slv{clingo}.

\section{A Novel Semantics for Declarative Domain-Specific Heuristics}
\label{sec-novel-semantics}

Supporting the declarative specification of domain-specific heuristics in ASP plays an important role in enabling ASP to solve large-scale industrial problems.
Although the language and semantics of heuristic directives in \slv{clingo} have shown to be beneficial in many cases, dynamic aspects of negation as failure and aggregates in heuristic conditions have not been addressed satisfactorily.
An alternative approach is necessary.

In this paper, we present a novel semantics for heuristic directives in ASP that improves this situation.
In this section, we assume that the underlying solver can assign to any atom one of three values: \emph{true} (denoted with $\sigT$), \emph{false} ($\sigF$), and \emph{must-be-true} ($\sigM$) (cf.\ \cite{DBLP:conf/lpnmr/Weinzierl17}).
For solvers that do not use the third truth value $\sigM$, the following definitions can be used without modification, the set of atoms assigned must-be-true will just be empty in this case.
\begin{definition}
\label{def-heuristic-directive}
Let $ha_i$ ($0\geq i \geq n$) be heuristic atoms of the form $\heusign_i~a_i$, where $\heusign_i \in \{ +, -, \heusignempty \}$ is a sign symbol and $a_i$ is an atom, and $w$ and $l$ are terms.
Then a \emph{heuristic directive} is of the form \cref{eq-alpha-heuristic-directive} (which is just one possible syntax).
\begin{align}
  &\heudirstmt ~~ \heuat_0 : \heuat_1, \dots, \heuat_k, \naf{\heuat_{k+1}}, \dots, \naf{\heuat_n}. \qquad [w@l]&
  \label{eq-alpha-heuristic-directive}
\end{align}
The heuristics' head is given by $\heuat_0$ and its condition by $\{ \heuat_1, \dots, \heuat_k$, $\naf{\heuat_{k+1}}$, \dots, $\naf{\heuat_n} \}$, which is similar to a rule body.
A heuristic directive must be safe, i.e., all variables occurring in it must also occur in $\{ \heuat_1, \dots, \heuat_k \}$.
\end{definition}

Among the syntactical differences between \cref{eq-clingo-heuristic-directive} and \cref{eq-alpha-heuristic-directive}, our proposal differs from \slv{clingo}'s in the following ways:
Each heuristic atom in the condition contains a sign symbol, which is either $+$ (strongly positive), $-$ (strongly negative) or $\heusignempty$ (the empty sign).
In the condition, sign symbols are used to provide a richer way of controlling when it is satisfied.
The interpretation of atoms together with sign symbols and default negation in the heuristic condition is summarized in \cref{tab-signs}, where \textit{yes} in a cell means that the literal given in the same row is satisfied under the assignment given in the same column, and \textit{no} means the contrary.
The sign $+$ is only relevant for solvers distinguishing between $\sigT$ and $\sigM$.
In the heuristic head, the sign symbol is used to determine the truth value to be chosen by the heuristics.
If $\heusign_0$ is $+$ or empty, the heuristics makes the solver guess $a_0$ to be true; if it is $-$, $a_0$ will be made false.
We do not use the modifier $m$.
Instead of weight $w$ and tie-breaking priority $p$, we use terms $w$ and $l$ denoting weight and level as familiar from optimize statements in ASP-Core-2 \cite{aspcore2} or weak constraints in DLV \cite{DBLP:journals/tocl/LeonePFEGPS06}.
Level is more important than weight, both default to $0$, and together they are called priority.

Heuristics under \slv{clingo}'s semantics can easily be represented in our framework by replacing \enquote{\nafsymbol} by \enquote{$-$} and by replacing weight and priority by appropriate values for weight and level.
The converse is presumed not to hold.

\begin{table}
	\centering
	\begin{tabular}{r|cccc}
		&	$\{ \sigF a, \sigM a, \sigT a \} \cap \assignment = \emptyset$	& $\sigF a \in \assignment$	& $\sigM a \in \assignment$	& $\sigT a \in \assignment$ \\
		\hline
		$a$			& no		& no		& yes		& yes \\
		$+a$		& no		& no		& no		& yes \\
		$-a$		& no		& yes		& no		& no \\
		$\naf{a}$	& yes		& yes		& no		& no \\
		$\naf{+a}$	& yes		& yes		& yes		& no \\
		$\naf{-a}$	& yes		& no		& yes		& yes \\
	\end{tabular}
	\caption{Satisfaction of literals in heuristic conditions wrt a partial assignment \assignment.}
	\label{tab-signs}
\end{table}

We now describe our semantics more formally,
using the following notations in the definitions below:
The function $\fheuat$ maps a heuristic atom $\heuat_i$ to $a_i$ by removing the sign.
The \emph{head} of a heuristic directive $\heudir$ of the form \cref{eq-alpha-heuristic-directive} is denoted by $\head(\heudir) = \heuat_0$, its \emph{weight} by $\mathrm{weight}(\heudir) = w$ if given, else 0, and its \emph{level} by $\mathrm{level}(\heudir) = l$ if given, else 0.
The \emph{(heuristic) condition} of a heuristic directive $\heudir$ is denoted by $\cond{\heudir} := \{ \heuat_1, \dots, \heuat_k,$ $\naf{\heuat_{k+1}}$, $\dots$, $\naf{\heuat_n} \}$,
the \emph{positive condition} is $\condpos{\heudir} := \{ \heuat_1, \dots, \heuat_k \}$ and the \emph{negative condition} is $\condneg{\heudir}$ $:=$ $\{ \heuat_{k+1}$, $\dots$, $\heuat_{n} \}$.

\begin{definition}
\label{def-condition-satisfied}
Given a ground heuristic directive $\heudir$ and a partial assignment $\assignment$, $\cond{\heudir}$ is \emph{satisfied} wrt $\assignment$ iff $\condpos{\heudir} \subseteq \assignmentpm$ and $\condneg{\heudir} \cap \assignmentpm = \emptyset$.
\end{definition}
Intuitively, a heuristic condition is satisfied if and only if its positive part is fully satisfied and none of its default-negated literals are contradicted.

\begin{definition}
\label{def-directive-applicable}
A ground heuristic directive $\heudir$ is \emph{applicable} wrt a partial assignment $\assignment$ iff:
$
\cond{\heudir}$ is satisfied, $\sigT \fheuat(\head(\heudir)) \notin \assignment$, and $\sigF \fheuat(\head(\heudir)) \notin \assignment
$.
\end{definition}
Intuitively, a heuristic directive is applicable if and only if its condition is satisfied and its head is assigned neither $\sigT$ nor $\sigF$.
If the head is $\sigM$, the heuristic directive may still be applicable,
because any atom with the non-final truth value $\sigM$ must be either $\sigT$ or $\sigF$ in any answer set.

\Cref{def-condition-satisfied,def-directive-applicable} reveal the main difference between the semantics proposed here and the one implemented by \slv{clingo}:
In our approach, $+$ and $-$ signs can be used in heuristic conditions to reason about atoms that are already assigned $\sigT$ or $\sigF$ in a partial assignment, while default negation can be used to reason about atoms that are assigned \emph{or still unassigned}.
Our semantics truly means default negation in the current partial assignment, while the one implemented by \slv{clingo} basically amounts to strong negation in the current search state.
This difference is crucial, since reasoning about incomplete information is important in many cases.
An example is a heuristics for bin packing that only applies to items not yet placed.

What remains to be defined is the semantics of weight and level.
Given a set of applicable heuristic directives, from the ones on the highest level one with the highest weight will be chosen.
If there are several with the same maximum priority, the solver can use a domain-independent heuristics like VSIDS \cite{DBLP:conf/dac/MoskewiczMZZM01} as a fallback to break the tie.
\begin{definition}
\label{def-maxpriority}
Given a set $\heudirs$ of applicable ground heuristic directives, the \emph{subset eligible for immediate choice} is defined as $\mathrm{maxpriority}(\heudirs)$ in two steps:
\begin{align*}
\mathrm{maxlevel}(\heudirs) &:= \{ \heudir \mid \heudir \in \heudirs ~\mathrm{and}~ \mathrm{level}(\heudir) = \max_{\heudir \in \heudirs}~\mathrm{level}(\heudir) \} \\
\mathrm{maxpriority}(\heudirs) &:= \{ \heudir \mid \heudir \in \mathrm{maxlevel}(\heudirs) ~\mathrm{and}~ \mathrm{weight}(\heudir) = \max_{\heudir \in \mathrm{maxlevel}(\heudirs)} \mathrm{weight}(\heudir) \}
\end{align*}
\end{definition}

After choosing a heuristics using $\mathrm{maxpriority}$, a
solver makes a decision on the directive's head atom.
Note that heuristics only choose between atoms derivable by currently applicable rules.
Other solving procedures, e.g., deterministic propagation, are unaffected by this.

\begin{example}
Consider the program given in \cref{sec-soa}.
Its heuristic directives, when converted to the syntax proposed in \cref{def-heuristic-directive}, look like directives \cref{dir1,dir2} in the following program.
Consider also the newly introduced directives \cref{dir3,dir4} in this program.

\noindent\begin{minipage}{.45\linewidth}
	\begin{align}
	&\heudirstmt ~~ \mathrm{a(5)}.   &[1]									\label{dir1}\\
	&\heudirstmt ~~ \mathrm{a(4)} ~~ : ~~ \naf{\mathrm{a(5)}}. &[2]	\label{dir2}
	\end{align}
\end{minipage}%
\hfill
\begin{minipage}{.50\linewidth}
	\begin{align}
	&\heudirstmt ~~ \mathrm{-a(5)} ~~ : ~~ \mathrm{a}(4). &[2]			\label{dir3}\\
	&\heudirstmt ~~ \mathrm{a(6)} ~~ : ~~ \mathrm{-a}(5), \mathrm{+a}(4). &[2]			\label{dir4}
	\end{align}
\end{minipage}
\vspace{1em}

Intuitively, directive \cref{dir1} unconditionally prefers to make $\mathrm{a(5)}$ true with weight 1.
All other directives have a higher weight, 2, but they become applicable at different points in time.
Directive \cref{dir2} prefers to make $\mathrm{a(4)}$ true if $\mathrm{a(5)}$ is not true, directive \cref{dir3} prefers to make $\mathrm{a(5)}$ false if $\mathrm{a(4)}$ is true or must-be-true, and \cref{dir4} prefers to make $\mathrm{a(6)}$ true if $\mathrm{a(5)}$ is false and $\mathrm{a(4)}$ is true.

Let $\assignmentpm_0 = \emptyset$ be the signed atom projection of the empty partial assignment before any decision has been made.
Wrt $\assignmentpm_0$, \cref{dir1} is applicable because its condition is empty and its head is still unassigned.
Directive \cref{dir2} is also applicable, because $\mathrm{a}(5)$ is still unassigned.
Directives \cref{dir3,dir4} are not applicable wrt $\assignmentpm_0$.
\cref{dir2} is chosen because it has the higher priority.
Thus, $\mathrm{a}(4)$ is assigned $\sigT$, which updates our signed atom projection to $\assignmentpm_1 = \{ \mathrm{a(4), +a(4)} \}$.
This makes \cref{dir3} applicable, $\mathrm{a}(5)$ is assigned $\sigF$ and our projection is $\assignmentpm_2 = \{ \mathrm{a(4), +a(4), -a(5)} \}$.
Note that the condition of \cref{dir2} was still satisfied at this point, but it was not applicable because its head was already assigned.
Now, also \cref{dir1} is not applicable anymore and the only directive that remains is \cref{dir4}.
Since it is applicable, $\mathrm{a(6)}$ is made true and added to the assignment.
Next, the atoms that remained unassigned are guessed by the default heuristics until an answer set is found.
\end{example}

\begin{example}
\label{ex-bpp-alpha}
Continuing from \cref{ex-bpp-clingo}, one possible encoding of the best-fit heuristics for bin packing (BPP) involving a heuristic directive of the form given in \cref{def-heuristic-directive} looks as follows:
\begin{align*}
\heudirstmt ~ \mathrm{in}(I,B) :~ &\mathrm{bin}(B), \mathrm{item}(I), \mathrm{bcap}(C), C \geq F+I, \naf{+\mathrm{item\_placed}(I)},\\
	&\mathrm{filled\_at\_least}(B,F), \naf{\mathrm{filled\_at\_least}(B,F+1)}. ~ [F+I] \\
\mathrm{filled\_at\_least}(B,F) \leftarrow~ &\mathrm{bin}(B), \mathrm{possible\_fill\_degree}(F), F \leq \sumagg ~ \{ ~ I' ~ : \mathrm{in}(I',B) ~ \}. \\
\mathrm{possible\_fill\_degree}(0..C) \leftarrow~ &\mathrm{bcap}(C). \\
\mathrm{item\_placed}(I) \leftarrow~ &\mathrm{in}(I,\_).
\end{align*}
Through the condition $\naf{+\mathrm{item\_placed}(I)}$, only those items are attempted to be placed that have not been placed yet.
Here, it is important to use the $\mathrm{+}$ sign to avoid that the condition is accidentally switched off in case $\mathrm{item\_placed}(I)$ is propagated to \sigM.
\end{example}

Here, it is possible to achieve the desired behavior by using an aggregate not directly in the condition but in a separate rule.
We leave the exact definition of the semantics of aggregates in heuristic conditions, which could involve heuristic atoms themselves, to future work.

\section{Integration into a Lazy-Grounding ASP Solver}
\label{sec-integration}

Most ASP systems split the evaluation into grounding and solving.
The former produces the grounding of a program, i.e., its variable-free equivalent
in which all variables are substituted by ground terms.
The latter then solves this propositional encoding.
The associated blow-up in space leads to the so-called \textit{grounding bottleneck} which is tackled by lazy grounding \cite{DBLP:conf/lpnmr/Weinzierl17,DBLP:conf/inap/LeutgebW17}.

The approach presented in \cref{sec-novel-semantics} is not tailored towards a specific solving paradigm in ASP.
We now describe how to integrate it into a lazy-grounding ASP solver.
Integration into a ground-and-solve system belongs to future work.
The system we are working with is \alphaslv\ \cite{DBLP:conf/lpnmr/Weinzierl17}, which is briefly described in the following paragraphs.
Note that \alphaslv\ does not (yet) support the full language of ASP,
e.g., it lacks upper bounds of cardinality and aggregate atoms.

\alphaslv, whose source code is freely available,\footnote{\alphaslv\ sources and binaries can be found on \url{https://github.com/alpha-asp/Alpha}. Features described in this section have been implemented on the \texttt{domspec_heuristics} branch and will soon be merged to \texttt{master}.} combines lazy grounding with Conflict-Driven Nogood Learning (CDNL) search (cf.\ \cite{DBLP:journals/ai/GebserKS12}) to avoid the
grounding bottleneck of ASP and obtain very good search performance.
CDNL-based ASP solvers require a fully grounded input, usually in the form of
nogoods. \alphaslv{} provides this by having two dedicated components, a lazy grounder and a
modified CDNL solver, as
is common for pre-grounding ASP solvers. For \alphaslv{} these
components interact cyclically: whenever the solver derives new
truth assignments to atoms, the grounder is queried for new ground
nogoods obtainable by the new assignments. In contrast to traditional CDNL-based solving,
this interplay results in a computation sequence.

Most importantly, the solver does not guess on each atom whether it is true or
false, but it guesses on ground instances of rules whether they fire or not.
This is realised by creating a unique atom for each ground body and then guessing on these body-representing atoms (a.k.a.\ \emph{choice points}).
The solver can choose a choice point if it is \emph{active}, i.e., when the corresponding ground rule is \emph{applicable}.
A rule is applicable w.r.t a partial assignment $\assignmentpm$ if its positive body has already been derived and its negative body is not contradicted, i.e., $\bodyp(r) \subseteq \assignmentpm$ and $\bodyn(r) \cap \assignmentpm = \emptyset$.
Intuitively, the \alphaslv{} algorithm incrementally grounds those rules whose positive body is already satisfied.
Supportedness of answer sets is not achieved by completion nogoods,
but by special nogoods with heads and the notion of justified truth \cite{DBLP:conf/lpnmr/Weinzierl17,DBLP:conf/ijcai/BogaertsW18}.

To be able to reuse standard grounding procedures, a preprocessor component in \alphaslv{} transforms heuristic directives occurring in input programs to normal rules with head atoms of a built-in predicate, henceforth called \emph{heuristic rules}.
The body of the heuristic rule equals the heuristic condition, while information on weight, level, heuristic head and sign are stored in the head of the rule.
Due to this transformation, a heuristic directive is grounded under the same precondition as a normal rule: which is, when its positive body is satisfied.

The way heuristic directives are grounded is similar to the way other rules are grounded, but still has to be different.
Bodies of heuristic rules are not represented as choice points, but as a distinct type of atoms, s.t.\ the solver can treat them differently.
When a heuristic rule is applicable (i.e., its positive body is derived and its negative body is not contradicted), it is not eligible for choice but the heuristics itself becomes applicable.
When it ceases to be applicable, the corresponding heuristic information is also not used by the solver any longer.
Thus, a heuristic condition is satisfied if and only if its corresponding heuristic rule is applicable.

The task of finding the applicable heuristics with the highest priority is aided by the use of efficient data structures like a heap.
When this heuristics is found, it cannot be chosen immediately in a lazy-grounding system like \alphaslv{}.
The reason for this is that the head of a heuristic directive is an ordinary atom, but \alphaslv{} can only choose on choice points -- on atoms representing rule bodies.
Therefore, an additional step is necessary:
The set of known ground rules that can derive the chosen atom is determined.
From this set, a fallback heuristics chooses one.
Then, the choice point corresponding to this rule is assigned true or false, depending on the sign given by the heuristic directive.
Propagation following this choice will immediately assign the desired truth value to the atom originally chosen.\footnote{If the head of the heuristic is negative and there are several rules deriving the atom in the head, setting one of them to false does not guarantee that the atom will propagate to false.}

\begin{example}
Consider the following program $\prog$:
\vspace{0.2ex}
\begin{center}
	\begin{tabular}{l}
		$\mathrm{x}(1..2). \qquad \{ \mathrm{a}(X) : \mathrm{x}(X) \}.$\\
		$\mathrm{b}(X) \leftarrow \mathrm{x}(X), \naf{\mathrm{c}(X)}. \qquad \mathrm{c}(X) \leftarrow \mathrm{x}(X), \naf{\mathrm{b}(X)}.$\\
		$\heudirstmt\ \mathrm{b}(X) : \mathrm{x}(X), \naf{\mathrm{a}(X)}. ~~~ [X@2]$
	\end{tabular}
\end{center}
\vspace{0.2ex}
Let $\heupred/4$ be the built-in predicate used to define heads of heuristic rules.
In a preprocessing step, the heuristic statement in $\prog$ is first translated to the heuristic rule $\heupred(\mathrm{b}(X),X,2,\mathrm{true}) \leftarrow \mathrm{x}(X), \naf{\mathrm{a}(X)}.$
Since the positive body of every rule in $\prog$ is satisfied, the full grounding of $\prog$ is immediately produced.
Under the initial partial assignment consisting just of facts $\assignment = \{ \sigT \mathrm{x(1)}, \sigT \mathrm{x(2)} \}$, both ground heuristic rules are applicable, since both their positive bodies are satisfied and neither $\mathrm{a}(1)$ nor $\mathrm{a}(2)$ is assigned yet.
The directive in which $X$ has been substituted by $2$ has the higher weight, however.
For this reason, it is chosen, and the solver finds the (in this case) only rule that can make the heuristics' head $\mathrm{b}(2)$ true: $\mathrm{b}(2) \leftarrow \mathrm{x}(2), \naf{\mathrm{c}(2)}.$
The choice point representing the body of this rule is made true and, after some propagation, the new partial assignment will contain $\mathrm{b}(2)$ (amongst other consequences of propagation).
\end{example}

\section{Experimental Results}
\label{sec-experiments}

To demonstrate the feasibility of our approach, we ran a set of experiments on encodings of the House Reconfiguration Problem (HRP) and the Partner Units Problem (PUP).
We concentrate on problem instances whose grounding is very large, but which are easy to solve by specialized domain-specific heuristics.\footnote{Instances have been selected by first defining an instance-generating algorithm and then exploring instance sizes to find a range with large groundings and solving times well distributed within the time limit of 60 minutes.}
By this, we extend the application area of ASP because instances of such sizes cannot be solved by conventional ASP systems.
Unfortunately \alphaslv\ cannot yet be used to solve very hard problems in acceptable time because several techniques used by other solvers to speed up solving have not yet been implemented.
Almost none of the problem instances used in our experiments can be solved by \alphaslv\ without domain-specific heuristics or by \slv{clingo} when some techniques that are not supported by \alphaslv\ are switched off.\footnote{Such features can be switched off in \slv{clingo} by arguments \texttt{--sat-prepro=no --eq=0 -r no -d no}. In this configuration, \slv{clingo} could solve one out of 29 HRP instance and 9 out of 141 PUP instances. \alphaslv\ without domain-specific heuristics could solve 0 HRP instances and 7 PUP instances.}

Both HRP \cite{DBLP:conf/confws/FriedrichRFHSS11} and PUP \cite{teppan2016solving,DBLP:conf/cpaior/AschingerDFGJRT11} are abstracted versions of industrial (re)configuration problems.
For problem definitions, we refer to the original sources.
All encodings (including specifications of the heuristics) and instances as well as the \alphaslv\ binaries used for our experiments are available on our website.\footnote{\url{http://ainf.aau.at/dynacon}}
HRP has been used without optimisation statements, since \alphaslv\ does not support them yet.
However, heuristic directives can be written in a way that optimal or near-optimal solutions are preferably found.\footnote{The optimality of solutions can sometimes be assessed by comparing the value of the objective funktion, but for some problems the optimum is unknown.}
As mentioned above, instances have been chosen to be very large but easily solvable by domain-specific heuristics.
For HRP we used instances with empty legacy configurations (i.e., configuration, not reconfiguration problems).
HRP instance sizes ranged from 100 to 800 things and PUP instances from 10 to 150 units.

Heuristic encodings for \slv{clingo} have also been created.
The one for HRP contains heuristic directives that have \enquote{faithfully} been adapted: by
using sign modifiers instead of sign symbols in heuristic heads;
by adding $(l-1)$ times the maximum weight from the next lower level to $w$ and omitting $p$;
and by removing all literals from the condition for which it only makes sense to evaluate them wrt a partial assignment.
The one for PUP has been created in an effort to create a QuickPup*-like heuristics similar to the one created for \alphaslv.

It has been observed that permissive grounding (cf.\ \cite{DBLP:conf/lpnmr/TaupeWF19}), e.g., providing the solver with more nogoods representing ground rules than necessary, can be counterproductive when domain-specific heuristics are used, because a good domain-specific heuristics can assist the solver even better while avoiding the overhead of additional nogoods.
Therefore, \alphaslv{} was used in its default configuration.
The JVM
was called with command-line parameters \texttt{-Xms1G -Xmx32G}.
For comparison, \slv{clingo} \cite{DBLP:journals/corr/GebserKKS14} was used in version 5.3.0 and \slv{dlv2} \cite{DBLP:conf/lpnmr/AlvianoCDFLPRVZ17} in version 2.0.

Experiments were run on a cluster of machines each with two
Intel\textsuperscript{\textregistered} Xeon\textsuperscript{\textregistered} CPU E5-2650 v4 @ 2.20GHz with 12 cores each, 252 GB of memory, and Ubuntu 16.04.1 LTS Linux.
Benchmarks were scheduled with the ABC Benchmarking System \cite{DBLP:conf/aiia/Redl16} together with HTCondor\textsuperscript{\texttrademark}.\footnote{\url{https://github.com/credl/abcbenchmarking}, \url{http://research.cs.wisc.edu/htcondor}}
Time and memory consumption were measured by \slv{pyrunlim},\footnote{\url{https://alviano.com/software/pyrunlim/}}
which was also used to limit time consumption to 60 minutes per instance, memory to 40 GiB and swapping to 0.
Care was taken to avoid side effects between CPUs, e.g., by requesting two cores for each benchmark from HTCondor and by setting process niceness to -20.
For HRP and PUP one instance of each size was available, on which each solver was run seven times to report median results.

For each problem instance, all solvers have searched for the first answer set.
The rationale behind this is that, since computation of solutions for large instances can be quite challenging, it is often sufficient to find one or only a few solutions in industrial use cases \cite{DBLP:journals/aim/FalknerFHSS16}.
Therefore, the domain-specific heuristics used in the experiments are designed to assist the solver especially in finding one answer set that is \enquote{good enough}, even though it may not be optimal.

\begin{figure}[t]
	\centering
	\begin{subfigure}{.48\textwidth}
		\centering
		\includegraphics[height=4.1cm]{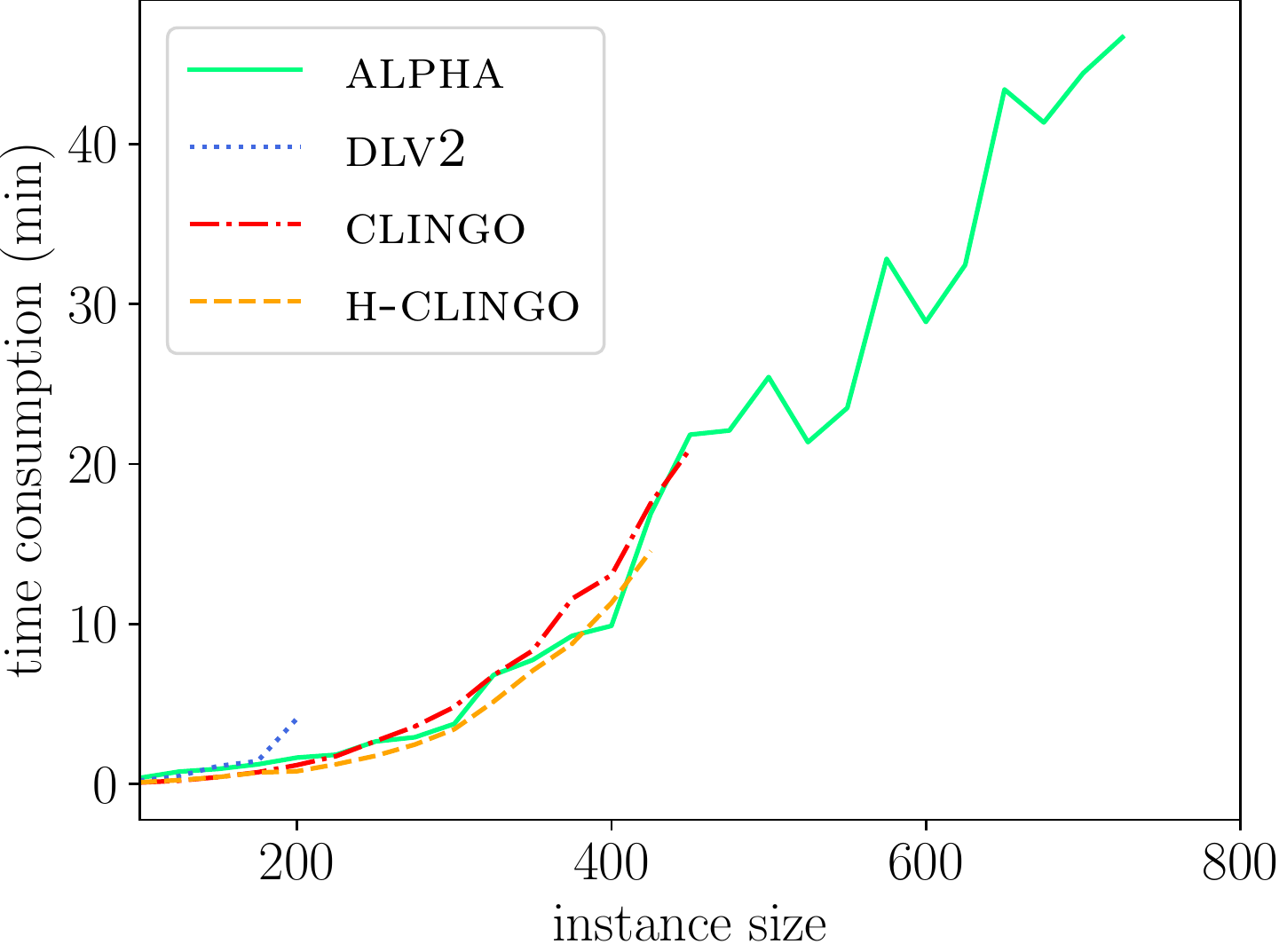}
		\caption{Time consumption in minutes}
		\label{fig-line-house-time}
	\end{subfigure}%
	\hspace*{\fill}
	\begin{subfigure}{.48\textwidth}
		\centering
		\includegraphics[height=4.1cm]{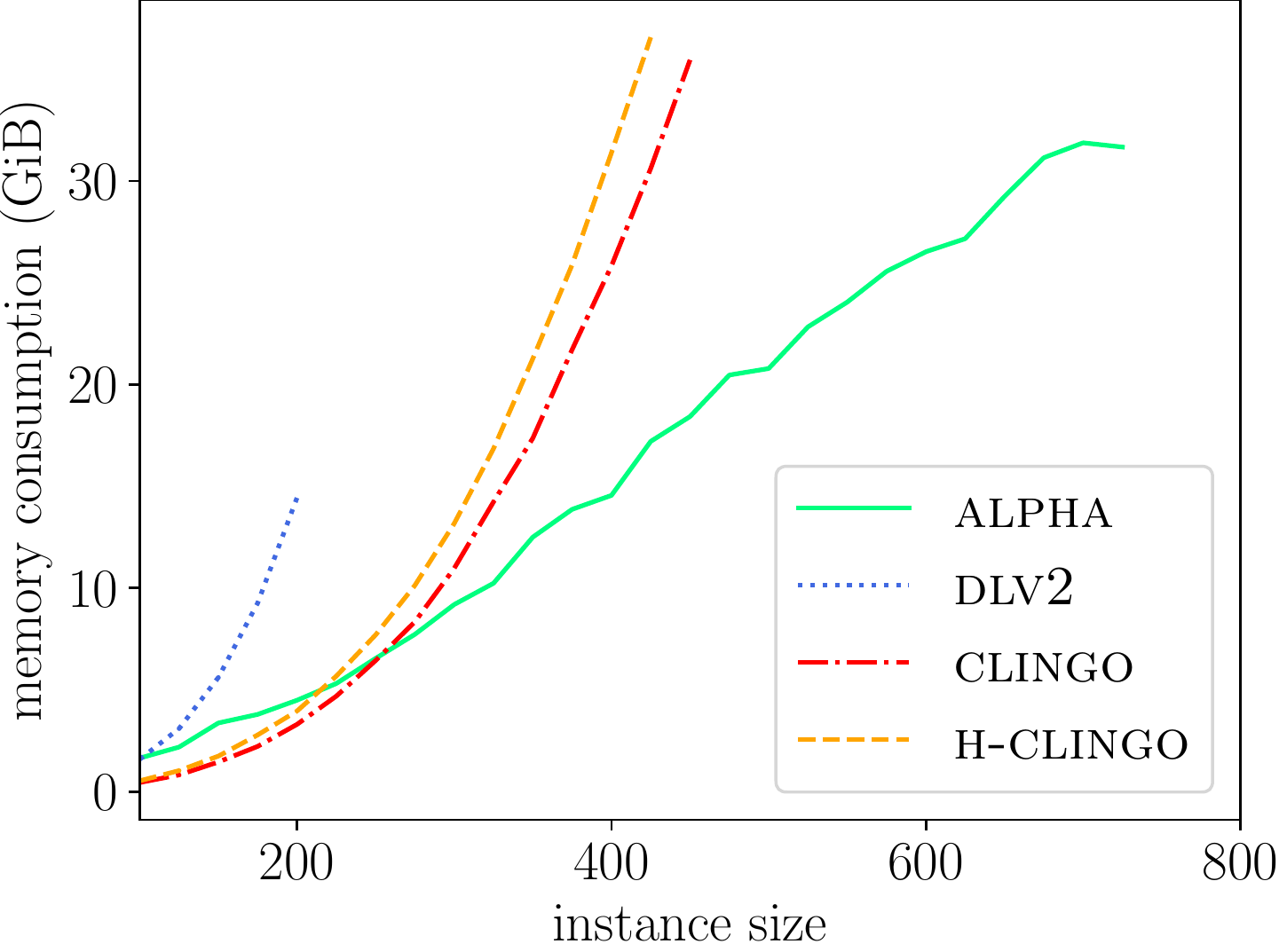}
		\caption{Memory consumption in GiB}
		\label{fig-line-house-memory}
	\end{subfigure}
	\caption{Time and memory consumption for finding the first answer set for HRP instances}
	\label{fig-experimental-results-hrp}
\end{figure}

Results for HRP are shown in \cref{fig-experimental-results-hrp}, consisting of time and memory consumption data for all three solvers.
\slv{clingo} is used both with (\slv{h-clingo}) and without domain-specific heuristics.
Domain-specific heuristics for \slv{dlv2} are out of scope because this system does not support the declarative specification of such heuristics.
A striking feature of these plots is that \slv{clingo} and \slv{dlv2} cannot solve most of the larger instances
within available memory, while \alphaslv{} solves all instances up to size 700 by successful application of lazy grounding.
Time consumption of \alphaslv{} with domain-specific heuristics compares well with \slv{clingo}'s.
When \slv{clingo} is equipped with a \enquote{faithful} adaption of the domain-specific heuristics as described above, it gets a bit faster, but solves fewer instances due to higher memory consumption.
Presumably, \slv{clingo} does not profit from domain-specific heuristics so much because its domain-independent heuristics and propagation are already very strong on this problem.
\slv{dlv2} exceeds available memory very quickly.

\begin{figure}[t]
\centering
\begin{subfigure}{.48\textwidth}
	\centering
	\includegraphics[height=4.1cm]{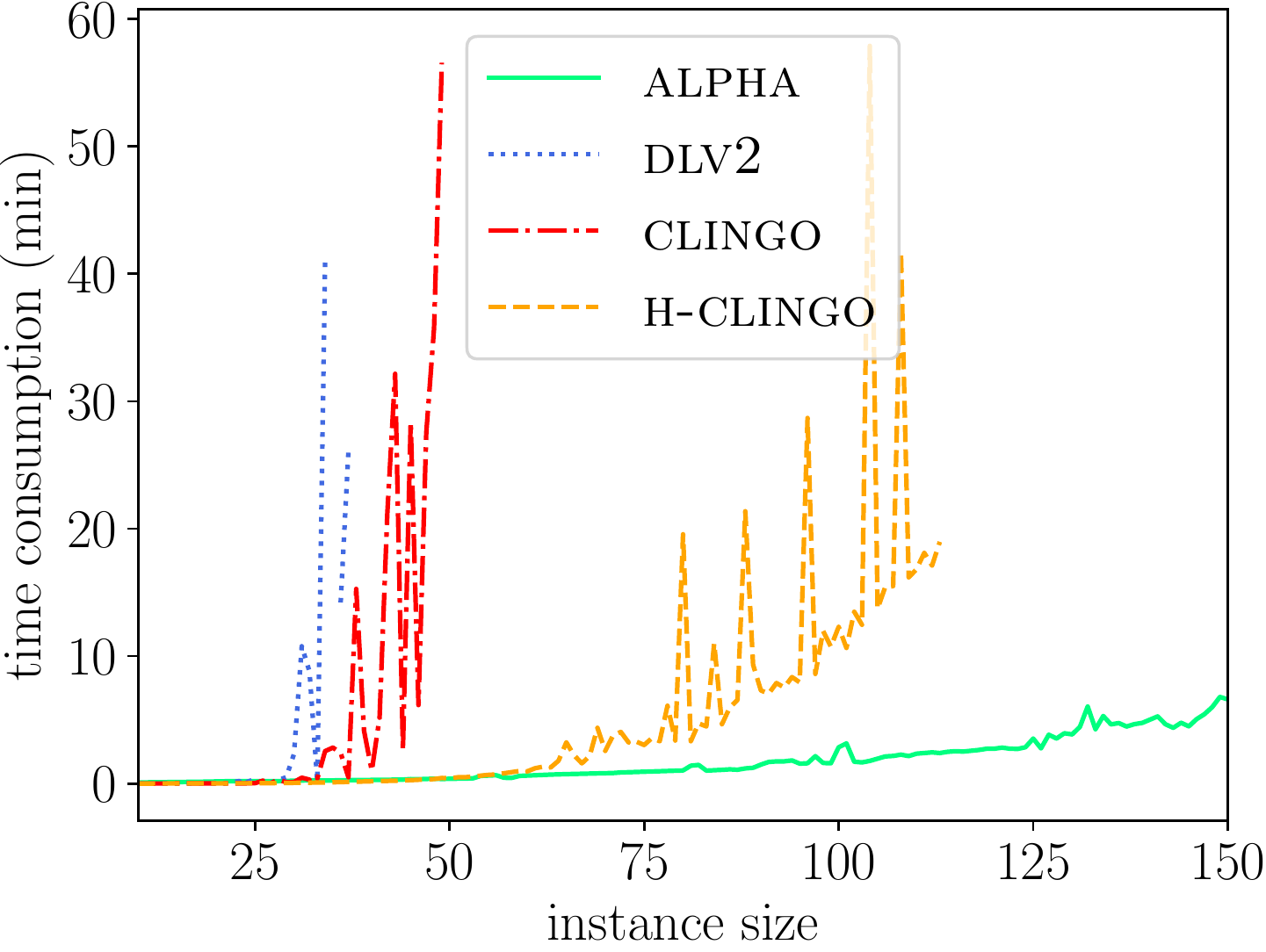}
	\caption{Time consumption in minutes}
	\label{fig-line-pup-time}
\end{subfigure}%
\hspace*{\fill}
\begin{subfigure}{.48\textwidth}
	\centering
	\includegraphics[height=4.1cm]{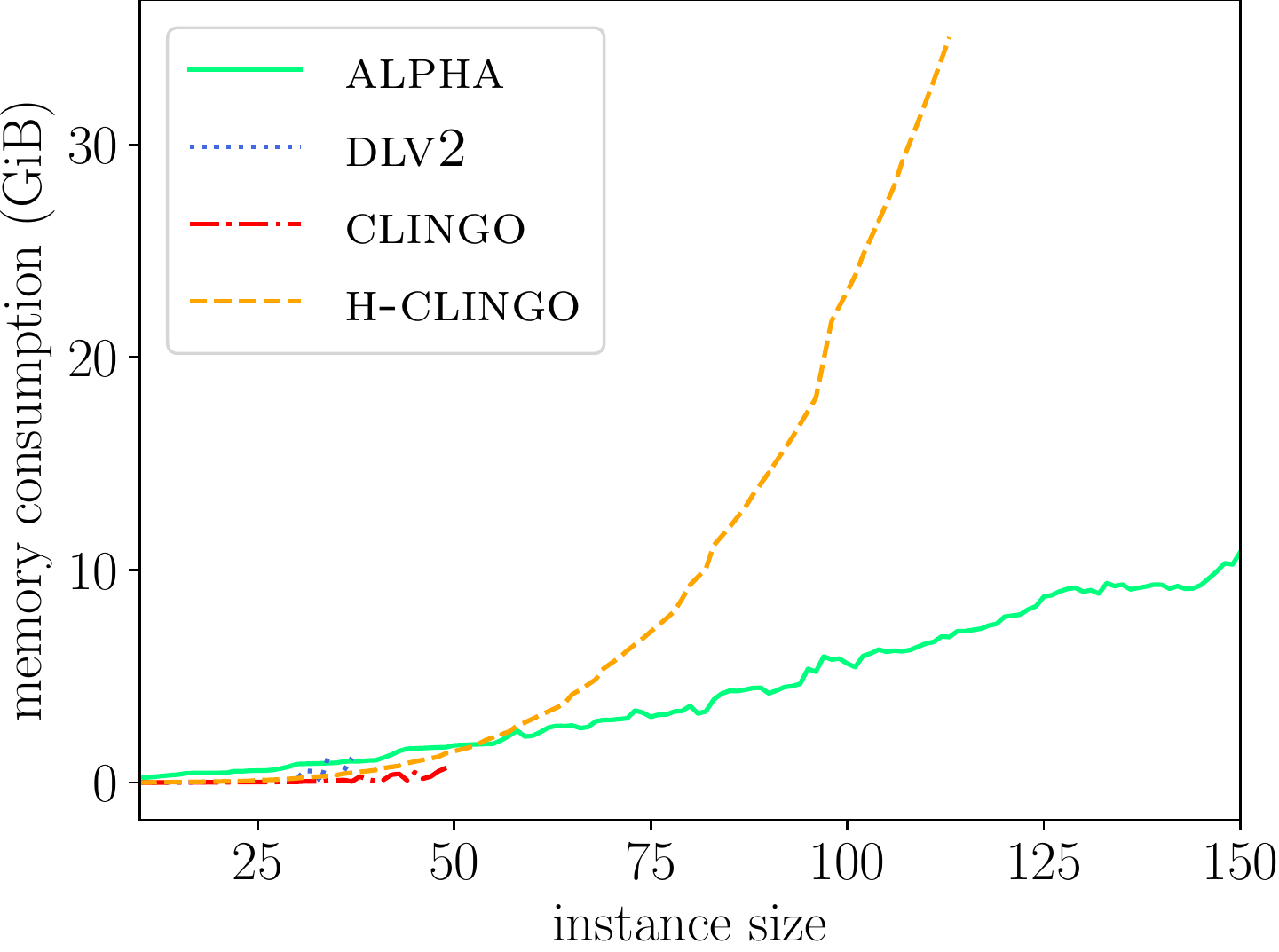}
	\caption{Memory consumption in GiB}
	\label{fig-line-pup-memory}
\end{subfigure}
\caption{Time and memory consumption for finding the first answer set for PUP instances}
\label{fig-experimental-results-pup}
\end{figure}

Results for PUP are shown in \cref{fig-experimental-results-pup}.
Only \alphaslv\ is able to solve all available instances within 60 minutes.
\alphaslv\ is able to achieve this with less memory and in less time than \slv{h-clingo} on the long run.
Here, \slv{dlv2} runs into time-outs early on.

Bin Packing (BPP) was also experimented with.
While the best-fit heuristics for this problem serves as a good example (cf.\ \cref{ex-bpp-alpha}), BPP is easily solvable by state-of-the-art solvers without domain-specific heuristics.
Experimental results (not shown due to lack of space) confirm our expectation that solvers do not benefit from domain-specific heuristics in this case.

\slv{clingo}-like heuristics can be approximated in \alphaslv\ by replacing \enquote{\nafsymbol} by \enquote{$-$}.
Cursory experiments with such encodings suggest that due to the lack of heuristic conditions exploiting negation as failure to avoid conflicting assignments, \alphaslv\ produces many backtracks and therefore uses much more time and space than with heuristics under the proposed semantics.

\section{Conclusions and Future Work}
\label{sec-conclusion}

We have proposed a novel semantics for declarative domain-specific heuristics in ASP, demonstrated how to integrate them in a lazy-grounding ASP system and presented experimental results obtained with the lazy-grounding solver \alphaslv.
Our semantics differs from the previous state of the art by evaluating default negation and aggregates with respect to incomplete information more naturally during solving.
Benefits of this semantics for many problem domains have been made evident by use of examples.
Our experimental results show that our approach is feasible, exhibiting encouraging time consumption and outstanding memory consumption behaviour.
Still, \alphaslv\ is far less mature than other ASP solvers and has to be improved in the future.
Techniques like restarts and nogood forgetting will further improve its performance.

Further work is also planned to extend syntax and semantics proposed here.
It could be worthwhile to adopt ideas like \texttt{init} and \texttt{factor} modifiers from \slv{clingo}.
Randomness and restarts should also be supported, since such features are used by several real-world domain-specific heuristics.
Since our approach has only been implemented in a lazy-grounding system so far, an adaption to ground-and-solve systems  like \slv{clingo} \cite{DBLP:journals/corr/GebserKKS14} or \slv{dlv2} \cite{DBLP:conf/lpnmr/AlvianoCDFLPRVZ17} should be investigated.

Thinking more broadly, the question how to generate domain-specific heuristics automatically is of great importance, since currently they have to be invented by humans familiar with the domain (and partly also with solving technology).

{
\footnotesize
\paragraph{Acknowledgements.}
This work has been conducted in the scope of the research project \textit{DynaCon (FFG-PNr.:\ 861263)}, which is funded by the Austrian Federal Ministry of Transport, Innovation and Technology (BMVIT) under the program \enquote{ICT of the Future} between 2017 and 2020.\footnote{See \url{https://iktderzukunft.at/en/} for more information.}
This research was also supported by the Academy of Finland, project 251170, by EU ECSEL Joint Undertaking under grant agreement no.\ 737459 (project Productive4.0),
and by the EU's Horizon 2020 research and innovation programme under grant agreement~825619 (AI4EU).
We also thank Andreas Falkner for his comments on an earlier version of this paper.
}

\bibliography{iclp}
\bibliographystyle{eptcsini}

\end{document}